\pdfoutput=1

\documentclass[11pt]{article}

\usepackage{acl}

\usepackage{times}
\usepackage{latexsym}

\usepackage[T1]{fontenc}

\usepackage[utf8]{inputenc}

\usepackage{microtype}

\usepackage{amsmath}
\usepackage{booktabs}
\usepackage{graphicx}
\usepackage[whole]{bxcjkjatype}
\usepackage{cleveref}
\usepackage{siunitx}

\newcommand{\scorewithinterval}[2]{\begin{tabular}{@{}r@{}}#1\\[-1mm]\small{$\pm$#2}\end{tabular}}

\newcommand{\significantscorewithinterval}[2]{\phantom{\textsuperscript{*}}\begin{tabular}{@{}r@{}}#1\\[-1mm]\small{$\pm$#2}\end{tabular}\raisebox{0.5em}{\textsuperscript{*}}}

\newcommand{\highestscore}[1]{\textbf{#1}}

\newcommand{\head}[1]{\noindent\textbf{#1}\hspace{1ex}}

\title{LEIA: Facilitating Cross-lingual Knowledge Transfer in \\Language Models with Entity-based Data Augmentation}

\author{Ikuya Yamada \\
  Studio Ousia, RIKEN \\
  \texttt{ikuya@ousia.jp} \\\And
  Ryokan Ri \\
  LY Corporation, SB Intuitions \\
  \texttt{ryou0634@gmail.com}
}

\begin{document}
\maketitle
\begin{abstract}
Adapting English-based large language models (LLMs) to other languages has become increasingly popular due to the efficiency and potential of cross-lingual transfer.
However, existing language adaptation methods often overlook the benefits of cross-lingual supervision.
In this study, we introduce LEIA, a language adaptation tuning method that utilizes Wikipedia entity names aligned across languages.
This method involves augmenting the target language corpus with English entity names and training the model using left-to-right language modeling.
We assess LEIA on diverse question answering datasets using 7B-parameter LLMs, demonstrating significant performance gains across various non-English languages.\footnote{The source code is available at \url{https://github.com/leia-llm/leia}.}
\end{abstract}

\section{Introduction}
\label{sec:intro}

While large language models (LLMs) are emerging as foundational technology~\cite{NEURIPS2020_1457c0d6}, their data hungriness restricts their application to a few resource-rich languages, with English being the most dominant among them~\cite{joshi-etal-2020-state}.
A promising strategy to broaden their scope is language adaptation tuning~\cite{muller2022cedille,yong-etal-2023-bloom}, where an already-pretrained LLM is further trained on a corpus of a language of interest.
The underlying motivation is that the model can leverage the knowledge acquired during pretraining to the target language.

However, this typical approach overlooks the potential benefits of incorporating cross-lingual supervision.
Although language models can learn cross-lingual knowledge from a mix of monolingual corpora~\cite{conneau-etal-2020-emerging}, knowledge sharing between languages is limited, and significant performance gaps still exist between English and non-English languages~\cite{ahuja-etal-2023-mega,Etxaniz2023DoML,huang-etal-2023-languages}.

\begin{figure}[t]
  \centering
  \includegraphics[width=\columnwidth]{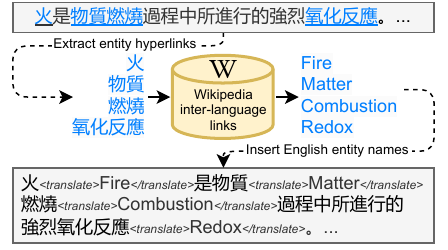}
  \caption{Data augmentation of LEIA applied to text from Chinese Wikipedia.
  English entity names, resolved through the inter-language links, enclosed in special {\small {\it <translate>}} and {\small {\it </translate>}} tokens are inserted adjacent to hyperlinks to facilitate cross-lingual transfer.}
  \label{fig:top_figure}
\end{figure}

In this work, we propose a language adaptation tuning method, LEIA ({\bf L}ightweight {\bf E}ntity-based {\bf I}nter-language {\bf A}daptation), that explicitly exploits cross-lingual supervision.
We focus on Wikipedia as a source of target language corpus, as it offers high-quality text data in a wide range of languages and the text contains hyperlinks to entities ({\it i.e.}, Wikipedia articles) that are aligned across different languages via inter-language links.
In our tuning phase, we insert an English entity name beside the corresponding entity in the text (see Figure \ref{fig:top_figure}), and train the model using the left-to-right language modeling objective.
This simple modification enables the model to extract and apply its English knowledge about the entities within the target language text during training, which we hypothesize to facilitate cross-lingual knowledge transfer.

We assess the effectiveness of LEIA through experiments using 7B-parameter LLMs, LLaMA 2~\citep{touvron2023llama} and Swallow~\cite{fuji-nakamura-etal-2024-swallow-llm}, and a diverse set of question answering datasets.
The results demonstrate that through our fine-tuning, LLMs benefit from knowledge transfer from English and significantly outperform the base models and those fine-tuned without LEIA.

\section{Method}
\label{sec:method}

Our method involves fine-tuning on a pretrained LLM using an augmented corpus derived from the target language edition of Wikipedia.
Specifically, for each hyperlink in the Wikipedia corpus, we insert the English name of the referred entity next to the hyperlink (Figure \ref{fig:top_figure}).
The English name is enclosed within special \textit{<translate>} and \textit{</translate>} tokens, allowing the model to identify its boundaries.
The English name of an entity is extracted from the title of the corresponding English Wikipedia page, which is identified using the inter-language links.
We ignore any hyperlinks pointing to entities not present in the English Wikipedia.\footnote{Across all the languages we experimented with, we successfully resolved over 80\% of the hyperlinks to their corresponding English Wikipedia pages using inter-language links.}
Further details are available in Appendix \ref{sec:detail-training}.

We fine-tune a pretrained LLM using the language modeling objective.
We train the model on the corpus of a target language and evaluate it using datasets in the same language.
The aforementioned special tokens are added to the vocabulary.
To prevent the model from generating these special tokens during inference, we block loss propagation when predicting these tokens during training.

\section{Experiments with LLaMA 2}
\label{sec:experiments_llama2}

We start with experiments using the LLaMA 2 7B model~\cite{touvron2023llama}.
Since it is primarily trained for English, it possesses a substantial amount of English knowledge that could be transferred to other languages.
Furthermore, its training corpus, containing approximately 38B non-English language tokens \cite{touvron2023llama}, fosters competitive multilingual performance~\cite{Etxaniz2023DoML}.
This makes LLaMA 2 a good candidate for investigating the effectiveness of our language adaptation method from English to other languages.

\subsection{Setup}
\label{sec:experiments_llama2_setup}

\head{Training}
We conduct experiments across seven languages: Arabic (ar), Spanish (es), Hindi (hi), Japanese (ja), Russian (ru), Swahili (sw), and Chinese (zh).
These languages are selected from five distinct language families (Appendix \ref{sec:languages}).
We fine-tune the model using up to 200 million tokens following \citet{yong-etal-2023-bloom}.
We use a batch size of 4 million tokens, following \citet{touvron2023llama}, resulting in 20 training steps for Swahili and 50 steps for other languages.\footnote{The fewer training steps for Swahili are due to the significantly smaller size of the Swahili Wikipedia corpus.}
The further details of the training are available in Appendix \ref{sec:detail-training}.

\head{Datasets}
We evaluate the model using two multiple-choice question answering datasets, X-CODAH and X-CSQA~\citep{lin-etal-2021-common}, which require commonsense knowledge to solve.
We present 0-shot results for X-CODAH and 4-shot results for X-CSQA.
Detailed information about these tasks is available in Appendix \ref{sec:detail-experiments}.

\head{Baselines}
As our primary baseline, we use a model fine-tuned under the same training settings as LEIA, using the original Wikipedia corpus without the insertion of English names (denoted as LLaMA2+FT).
Comparison with this baseline confirms that performance gains stem from the insertion of English names, not just from fine-tuning on the Wikipedia corpus.
We also use the random baseline and the LLaMA 2 model without fine-tuning.

\head{Method configurations}
We test three strategies to add the English name: (1) \textit{left}: inserting the name before the hyperlink, (2) \textit{right}: inserting the name after the hyperlink, and (3) \textit{replace}: replacing the original entity text with the name.
To reduce the train-test discrepancy, we randomly omit the insertion with a probability of $p_{\text{skip}}$.
The example in Figure \ref{fig:top_figure} adopts the \textit{right} strategy with $p_{\text{skip}}=0.0$.
Due to our limited computational resources, we test only $p_{\text{skip}} \in \{0.0, 0.5\}$.

\subsection{Results} 
\label{sec:llama2-results}

\begin{table}[t]
    \centering
    \scalebox{0.8}{
    \begin{tabular}{ccccc}
        \toprule
        strategy & $p_{\text{skip}}$ &  X-CODAH & X-CSQA \\
        \midrule
        left    & 0.0 &         35.6  &         30.5  \\
        left    & 0.5 & \textbf{36.1} & \textbf{30.6} \\
        \midrule
        right   & 0.0 &         35.8  &         30.5  \\
        right   & 0.5 & \textbf{36.1} & \textbf{30.6} \\
        \midrule
        replace & 0.0 &         35.8  &         30.4 \\
        replace & 0.5 &         36.0  &         30.5 \\
        \bottomrule
    \end{tabular}
    }
    \caption{Average accuracy scores across seven languages based on different method configurations. Full results are detailed in Table \ref{tab:full-scores-all-configurations}.}
    \label{tab:avg-scores-all-configurations}
\end{table}

We initially present the average accuracy across all languages for different method configurations in Table \ref{tab:avg-scores-all-configurations}. 
Overall, the choice of strategy has a minimal impact on performance.
Additionally, models with $p_{\text{skip}}=0.5$ consistently outperform their counterparts with $p_{\text{skip}}=0.0$ on both datasets.
To reduce computational costs, we exclusively use the \textit{right} strategy with $p_{\text{skip}}=0.5$ in subsequent experiments and refer to this setting as LEIA.

\begin{table*}[t]
    \centering
    \setlength{\tabcolsep}{0.61em}
    \scalebox{0.65}{
    \begin{tabular}{c|cccccccc|cccccccc}
        \toprule
        & \multicolumn{8}{c|}{X-CODAH} & \multicolumn{8}{c}{X-CSQA} \\
        Model & ar & es & hi & ja & ru & sw & zh & avg & ar & es & hi & ja & ru & sw & zh & avg \\
        \midrule
        Random &
          25.0 &
          25.0 &
          25.0 &
          25.0 &
          25.0 &
          25.0 &
          25.0 &
          25.0 &
          20.0 &
          20.0 &
          \highestscore{20.0} &
          20.0 &
          20.0 &
          \highestscore{20.0} &
          20.0 &
          20.0 \\
        LLaMA2 & 
          30.3 &
          45.3 &
          29.7 &
          30.3 &
          34.3 &
          28.7 &
          36.7 &
          33.6 &
          21.0 &
          45.1 &
          19.1 &
          34.4 &
          36.0 &
          16.0 &
          40.1 &
          30.2 \\	
        LLaMA2+FT &
          \scorewithinterval{30.7}{0.6} &
          \scorewithinterval{45.5}{0.4} &
          \scorewithinterval{27.2}{0.2} &
          \scorewithinterval{30.4}{0.3} &
          \scorewithinterval{34.4}{0.9} &
          \scorewithinterval{29.0}{0.1} &
          \scorewithinterval{38.3}{0.3} &
          33.6 &
          \scorewithinterval{21.3}{0.3} &
          \scorewithinterval{44.8}{0.2} &
          \scorewithinterval{18.2}{0.2} &
          \scorewithinterval{34.5}{0.3} &
          \scorewithinterval{35.7}{0.3} &
          \scorewithinterval{15.9}{0.1} &
          \scorewithinterval{39.7}{0.1} &
          30.0 \\
        \midrule
        LEIA & 
          \significantscorewithinterval{\highestscore{32.8}}{0.5} &
          \significantscorewithinterval{\highestscore{46.6}}{0.2} &
          \significantscorewithinterval{\highestscore{30.6}}{0.2} &
          \significantscorewithinterval{\highestscore{34.9}}{0.4} &
          \significantscorewithinterval{\highestscore{37.5}}{0.2} &
          \significantscorewithinterval{\highestscore{30.4}}{0.2} &
          \significantscorewithinterval{\highestscore{39.1}}{0.2} &
          \highestscore{36.0} &
          \significantscorewithinterval{\highestscore{21.9}}{0.2} &
          \significantscorewithinterval{\highestscore{45.7}}{0.1} &
          \scorewithinterval{18.4}{0.2} &
          \significantscorewithinterval{\highestscore{35.4}}{0.2} &
          \scorewithinterval{\highestscore{36.1}}{0.2} &
          \scorewithinterval{16.0}{0.1} &
          \significantscorewithinterval{\highestscore{40.5}}{0.1} &
          \highestscore{30.6}
          \\
        \bottomrule
    \end{tabular}
    }
    \caption{Results on X-CODAH and X-CSQA.
    For LLaMA2+FT and LEIA, we report mean accuracy and 95\% confidence intervals based on Student's t-distribution over 5 training runs with different random seeds.
    Scores of LEIA are marked with $^*$ if its improvement is statistically significant compared to all baselines.}
    \label{tab:codah-results}
\end{table*}

Table \ref{tab:codah-results} shows our main results.
LEIA outperforms all baseline models in all languages on X-CODAH and in 5 out of the 7 languages on X-CSQA.
Furthermore, LEIA outperforms the LLaMA2+FT baseline in all languages on both datasets. These results demonstrate that LEIA effectively enhances cross-lingual transfer.

Furthermore, all models, including LEIA, fail to surpass the random baseline in Hindi and Swahili on X-CSQA.
It appears that the models struggle to handle few-shot tasks in these two languages, likely due to the very limited presence of these languages in the pretraining corpora of LLaMA 2 \cite{touvron2023llama}.
Additionally, LLaMA2+FT does not outperform LLaMA 2 in several languages on both datasets.
We believe this decline could be due to Wikipedia's uniform, clean, and formal style.
Overfitting to this style might result in poor model performance on texts of different styles, such as casual, informal, and question-style texts.

Additional results based on different numbers of few-shot examples are available in Appendix \ref{sec:various-num-of-few-shots}.

\section{Experiments with Swallow}
\label{sec:experiments_swallow}

In this section, we examine if bilingual language models that already possess substantial knowledge not only in English but also in the target language can benefit from knowledge transfer from LEIA.
We focus on Japanese, a language with a variety of benchmark datasets, and experiment with the state-of-the-art English-Japanese LLM, Swallow 7B~\cite{fuji-nakamura-etal-2024-swallow-llm}.\footnote{\url{https://huggingface.co/tokyotech-llm/Swallow-7b-hf}}
This model was developed through continual pretraining on LLaMA 2 with vocabulary extension~\citep{cui2023efficient,nguyen2023seallms,zhao2024llama}, using bilingual corpora consisting of 90B Japanese tokens and 10B English tokens.
We show that even after the adaptation with a massive target language corpus, LEIA can further boost the performance of the model.

\subsection{Setup}

\head{Training}
We fine-tune the Swallow 7B model using the Japanese Wikipedia corpus, following the same training setup described in Section \ref{sec:experiments_llama2_setup}.

\head{Datasets}
In addition to X-CODAH and X-CSQA, we use four question answering datasets available in two tools for evaluating Japanese LLMs: JEMHopQA~\cite{ishii-jemhopqa} and NIILC~\cite{sekine-niilc} in llm-jp-eval~\cite{han-kiyomaru-etal-2024-llm-jp-eval},\footnote{
\url{https://github.com/llm-jp/llm-jp-eval}} and JCommonsenseQA~\cite{kurihara-etal-2022-jglue} and JAQKET~\cite{suzuki-jaqket} in the JP Language Model Evaluation Harness.\footnote{\url{https://github.com/Stability-AI/lm-evaluation-harness/tree/jp-stable}}
We present 4-shot results obtained with these tools. 
We use accuracy for JCommonsenseQA and JAQKET, and character-based F-measure for JEMHopQA and NIILC.
Further details are available in Appendix \ref{sec:detail-experiments}.

\head{Baselines}
We denote the model fine-tuned using the plain Wikipedia corpus as Swallow+FT.
We also evaluate Swallow without fine-tuning.

\subsection{Results}

\begin{table}[t]
    \centering
    \setlength{\tabcolsep}{0.35em}
    \scalebox{0.65}{
    \begin{tabular}{c|cccccc}
        \toprule
        Model & X-CODAH & X-CSQA & JCSQA & NIILC & JHQA & JAQKET \\
        \midrule
        Swallow &
          42.0 &
          41.0 &
          80.3 &
          59.5 &
          50.8 &
          39.1 \\
        Swallow+FT &
          \scorewithinterval{40.7}{0.3} & 
          \scorewithinterval{39.6}{0.2} & 
          \scorewithinterval{79.3}{0.1} & 
          \scorewithinterval{58.0}{0.3} & 
          \scorewithinterval{50.3}{0.8} & 
          \scorewithinterval{35.0}{0.8} \\
        \midrule
        LEIA & 
          \significantscorewithinterval{\highestscore{42.5}}{0.2} & 
          \significantscorewithinterval{\highestscore{42.1}}{0.1} & 
          \significantscorewithinterval{\highestscore{80.6}}{0.2} & 
          \significantscorewithinterval{\highestscore{60.3}}{0.2} & 
          \significantscorewithinterval{\highestscore{54.5}}{0.1} & 
          \significantscorewithinterval{\highestscore{41.3}}{0.6} \\
        \bottomrule

    \end{tabular}
    }
    \caption{Results on Japanese datasets.
    JCSQA and JHQA denote JCommonsenseQA and JEMHopQA, respectively.
    For fine-tuned models, we report mean accuracy and 95\% confidence intervals over 5 training runs.
    Scores of LEIA are marked with $^*$ if their improvement is statistically significant compared to all baselines.}
    \label{tab:swallow-results}
\end{table}

Table \ref{tab:swallow-results} shows that LEIA significantly outperforms all baseline models on all datasets.
This demonstrates the effectiveness of LEIA when the base model has already been trained with a massive corpus of the target language.
Furthermore, similar to the experimental results with LLaMA 2 (§\ref{sec:llama2-results}), we observe the performance degradation of Swallow+FT compared to Swallow without fine-tuning.

\begin{table*}[t]
    \centering
    \scalebox{0.72}{
    \begin{tabular}{p{22em}|p{33em}}
        \toprule
        LLaMA2-FT's predictions & LEIA's predictions\\
        \midrule
        He is trying to boil the ocean. He loves the ocean. & He is trying to boil the ocean. He wants to accomplish something that is impossible. \\
        The Eiffel tower is in London, England. & The Eiffel tower is in Paris. \\
        I hear my phone ring. I turn up the volume. & I hear my phone ring. I answer it. \\
        I drink too much alcohol. I am in the fourth grade. & I drink too much alcohol. I might be an alcoholic. \\
        A person is at a food court. The person runs a marathon. & A person is at a food court. The person buys a sandwich.  \\
         \bottomrule
    \end{tabular}
    }
    \caption{Comparison of X-CODAH predictions by LLaMA2-FT and LEIA, both fine-tuned on Japanese Wikipedia.
    Only English versions are presented here; original Japanese sentences are shown in Table \ref{tab:qualitative-analysis-japanese}.}
    \label{tab:qualitative-analysis}
\end{table*}

\begin{table}[t]
    \centering
    \scalebox{0.8}{
    \begin{tabular}{ccc}
        \toprule
        & X-CODAH & X-CSQA \\
        \midrule
        \begin{tabular}{@{}c@{}}Enable loss propagation\\from English entity tokens\end{tabular} & \textbf{36.1} & \textbf{30.6} \\
        \begin{tabular}{@{}c@{}}Disable loss propagation\\from English entity tokens\end{tabular} & 36.0 & \textbf{30.6} \\
        \bottomrule
    \end{tabular}
    }
    \caption{Comparison of LEIA models with loss propagation enabled vs. disabled for tokens in English entity names. Average accuracy scores across seven languages are presented. Detailed results are available in Table \ref{tab:prevent-token-loss-full-results}.}
    \label{tab:prevent-token-loss}
\end{table}

\begin{table}[t]
    \centering
    \scalebox{0.8}{
    \begin{tabular}{ccc}
        \toprule
        & X-CODAH & X-CSQA \\
        \midrule
        w/ special tokens & \textbf{36.1} & \textbf{30.6} \\
        w/o special tokens & 33.3 & 30.2 \\
        \bottomrule
    \end{tabular}
    }
    \caption{Comparison of LEIA models with and without using special tokens during training. Average accuracy scores across seven languages are presented.}
    \label{tab:results-special-tokens}
\end{table}

\section{Analysis}
\label{sec:analysis}

\head{Qualitative analysis}
We present five random predictions of LEIA and LLaMA2+FT (§\ref{sec:experiments_llama2}) from the Japanese X-CODAH dataset, where LEIA answered correctly but the LLaMA2+FT failed in Table \ref{tab:qualitative-analysis}.
They demonstrate that LEIA effectively acquires commonsense knowledge ({\it e.g.,} the sea cannot be boiled) and factual knowledge ({\it e.g.,} the Eiffel Tower is in Paris) via cross-lingual knowledge transfer from English.

\head{How does LEIA facilitate transfer?}
The English names inserted into the corpus can enhance the training in two ways: (1) \textit{names as labels}: serving as labels to predict based on the preceding tokens, and (2) \textit{names as contexts}: providing context for the subsequent tokens.
Both aspects can facilitate cross-lingual transfer, allowing the model to apply knowledge from one language to another.
To determine which causes the performance improvements, we remove the effect of using \textit{names as labels} by blocking loss propagation when predicting tokens in English entity names.

The results in Table \ref{tab:prevent-token-loss} show that preventing loss propagation from the English entity tokens has a minimal impact on performance.
This indicates that LEIA's performance enhancement is mainly attributed to using \textit{names as contexts} in training.

\head{Effects of special tokens} To investigate the effects of the special \textit{<translate>} and \textit{</translate>} tokens during training, we conduct the training on LLaMA 2 without using these tokens when inserting English names.

The results in Table \ref{tab:results-special-tokens} show that performance consistently declines on both the X-CODAH and X-CSQA datasets when these special tokens are not used during training.
This suggests that these special tokens enable the model to identify the boundaries of inserted English names and play a crucial role in training.

\section{Related Work}
\label{sec:related_work}

\head{Language adaptation}
A common domain adaptation technique for language models is training on a domain-specific corpus~\citep{gururangan-etal-2020-dont}, and when different languages are considered as different domains, it can be used for language adaptation.
This strategy is shown to be effective in various models including encoder-decoder models~\citep{neubig-hu-2018-rapid}, bidirectional language models~\citep{han-eisenstein-2019-unsupervised,wang-etal-2020-extending,chau-etal-2020-parsing}, and auto-regressive language models~\citep{muller2022cedille,yong-etal-2023-bloom}.
However, when adapting to a new language, knowledge transfer from one language to another can be insufficient due to discrepancies in the surface forms.
To facilitate the sharing of internal knowledge across languages, our proposed method leverages cross-lingually aligned entity names.

\head{Cross-lingual supervision for language models}
To enhance cross-lingual transfer, incorporating cross-lingual supervision is effective. This supervision can come from various sources, including bilingual dictionaries and bitext~\cite{NEURIPS2019_c04c19c2,kale-etal-2021-nmt5,reid-artetxe-2022-paradise,wang-etal-2022-expanding}.
Wikipedia hyperlinks, which can be considered a special case of words or phrases aligned via a bilingual dictionary, have also been shown to be effective~\cite{XLM-K-2022-aaai,ri-etal-2022-mluke}.
Exploring the potential of Wikipedia is promising as its high-quality formal text and the continual expansion of data across many languages.
Our study showcases the benefits of using cross-lingually aligned entity names in continual training of language models.

\section{Conclusion}
\label{sec:conclusion}

We introduced LEIA, a method to facilitate cross-lingual knowledge transfer via fine-tuning language models using Wikipedia text augmented with English entity names.
We applied LEIA to the English LLM, LLaMA 2, and the English-Japanese LLM, Swallow, demonstrating significant improvements on various non-English question answering tasks.

Future research will investigate LEIA's effectiveness using an arbitrary text corpus with entity annotations generated through entity linking, instead of the Wikipedia corpus, and employing the augmented corpus during the pretraining of bi- or multi-lingual LLMs, rather than relying on post-hoc fine-tuning.

\section{Limitations}
\label{sec:limitation}

Our evaluation focused on question answering tasks, on the basis of the assumption that knowledge transferred from entity names mainly includes commonsense and world knowledge. While this type of knowledge has the potential to benefit a wider range of tasks, broader evaluation is left for future research.

The data source for our method is Wikipedia, limiting our coverage to languages represented therein.
However, our approach is adaptable to incorporate other forms of cross-lingual supervision such as bilingual dictionaries.
Such extensions could enhance the applicability of our proposed framework to additional languages not currently covered by Wikipedia.

\bibliography{reference}

\begin{thebibliography}{35}
\expandafter\ifx\csname natexlab\endcsname\relax\def\natexlab#1{#1}\fi

\bibitem[{Ahuja et~al.(2023)Ahuja, Diddee, Hada, Ochieng, Ramesh, Jain, Nambi, Ganu, Segal, Ahmed, Bali, and Sitaram}]{ahuja-etal-2023-mega}
Kabir Ahuja, Harshita Diddee, Rishav Hada, Millicent Ochieng, Krithika Ramesh, Prachi Jain, Akshay Nambi, Tanuja Ganu, Sameer Segal, Mohamed Ahmed, Kalika Bali, and Sunayana Sitaram. 2023.
\newblock \href {https://doi.org/10.18653/v1/2023.emnlp-main.258} {{MEGA}: Multilingual evaluation of generative {AI}}.
\newblock In \emph{Proceedings of the 2023 Conference on Empirical Methods in Natural Language Processing}, pages 4232--4267, Singapore. Association for Computational Linguistics.

\bibitem[{Brown et~al.(2020)Brown, Mann, Ryder, Subbiah, Kaplan, Dhariwal, Neelakantan, Shyam, Sastry, Askell, Agarwal, Herbert-Voss, Krueger, Henighan, Child, Ramesh, Ziegler, Wu, Winter, Hesse, Chen, Sigler, Litwin, Gray, Chess, Clark, Berner, McCandlish, Radford, Sutskever, and Amodei}]{NEURIPS2020_1457c0d6}
Tom Brown, Benjamin Mann, Nick Ryder, Melanie Subbiah, Jared~D Kaplan, Prafulla Dhariwal, Arvind Neelakantan, Pranav Shyam, Girish Sastry, Amanda Askell, Sandhini Agarwal, Ariel Herbert-Voss, Gretchen Krueger, Tom Henighan, Rewon Child, Aditya Ramesh, Daniel Ziegler, Jeffrey Wu, Clemens Winter, Chris Hesse, Mark Chen, Eric Sigler, Mateusz Litwin, Scott Gray, Benjamin Chess, Jack Clark, Christopher Berner, Sam McCandlish, Alec Radford, Ilya Sutskever, and Dario Amodei. 2020.
\newblock \href {https://proceedings.neurips.cc/paper_files/paper/2020/file/1457c0d6bfcb4967418bfb8ac142f64a-Paper.pdf} {Language models are few-shot learners}.
\newblock In \emph{Advances in Neural Information Processing Systems}, volume~33, pages 1877--1901. Curran Associates, Inc.

\bibitem[{Chau et~al.(2020)Chau, Lin, and Smith}]{chau-etal-2020-parsing}
Ethan~C. Chau, Lucy~H. Lin, and Noah~A. Smith. 2020.
\newblock \href {https://doi.org/10.18653/v1/2020.findings-emnlp.118} {Parsing with multilingual {BERT}, a small corpus, and a small treebank}.
\newblock In \emph{Findings of the Association for Computational Linguistics: EMNLP 2020}, pages 1324--1334, Online. Association for Computational Linguistics.

\bibitem[{Chen et~al.(2019)Chen, D{'}Arcy, Liu, Fernandez, and Downey}]{chen-etal-2019-codah}
Michael Chen, Mike D{'}Arcy, Alisa Liu, Jared Fernandez, and Doug Downey. 2019.
\newblock \href {https://doi.org/10.18653/v1/W19-2008} {{CODAH}: An adversarially-authored question answering dataset for common sense}.
\newblock In \emph{Proceedings of the 3rd Workshop on Evaluating Vector Space Representations for {NLP}}, pages 63--69, Minneapolis, USA. Association for Computational Linguistics.

\bibitem[{Conneau and Lample(2019)}]{NEURIPS2019_c04c19c2}
Alexis Conneau and Guillaume Lample. 2019.
\newblock Cross-lingual language model pretraining.
\newblock In \emph{Proceedings of the Advances in Neural Information Processing Systems 32}, pages 7057--7067, Vancouver, BC, Canada.

\bibitem[{Conneau et~al.(2020)Conneau, Wu, Li, Zettlemoyer, and Stoyanov}]{conneau-etal-2020-emerging}
Alexis Conneau, Shijie Wu, Haoran Li, Luke Zettlemoyer, and Veselin Stoyanov. 2020.
\newblock \href {https://doi.org/10.18653/v1/2020.acl-main.536} {Emerging cross-lingual structure in pretrained language models}.
\newblock In \emph{Proceedings of the 58th Annual Meeting of the Association for Computational Linguistics}, pages 6022--6034, Online. Association for Computational Linguistics.

\bibitem[{Cui et~al.(2023)Cui, Yang, and Yao}]{cui2023efficient}
Yiming Cui, Ziqing Yang, and Xin Yao. 2023.
\newblock \href {https://api.semanticscholar.org/CorpusID:258180548} {Efficient and effective text encoding for {C}hinese {LLaMA} and {A}lpaca}.
\newblock \emph{ArXiv}, abs/2304.08177.

\bibitem[{Dao(2023)}]{Dao2023FlashAttention2FA}
Tri Dao. 2023.
\newblock \href {https://api.semanticscholar.org/CorpusID:259936734} {{FlashAttention-2}: Faster attention with better parallelism and work partitioning}.
\newblock \emph{ArXiv}, abs/2307.08691.

\bibitem[{Etxaniz et~al.(2023)Etxaniz, Azkune, Etxabe, de~Lacalle, and Artetxe}]{Etxaniz2023DoML}
Julen Etxaniz, Gorka Azkune, Aitor~Soroa Etxabe, Oier~Lopez de~Lacalle, and Mikel Artetxe. 2023.
\newblock \href {https://api.semanticscholar.org/CorpusID:260378999} {Do multilingual language models think better in english?}
\newblock \emph{ArXiv}, abs/2308.01223.

\bibitem[{Fujii et~al.(2024)Fujii, Nakamura, Loem, Iida, Ohi, Hattori, Shota, Mizuki, Yokota, and Okazaki}]{fuji-nakamura-etal-2024-swallow-llm}
Kazuki Fujii, Taishi Nakamura, Mengsay Loem, Hiroki Iida, Masanari Ohi, Kakeru Hattori, Hirai Shota, Sakae Mizuki, Rio Yokota, and Naoaki Okazaki. 2024.
\newblock \href {https://api.semanticscholar.org/CorpusID:269449465} {Continual pre-training for cross-lingual {LLM} adaptation: Enhancing {J}apanese language capabilities}.
\newblock \emph{ArXiv}, abs/2404.17790.

\bibitem[{Gururangan et~al.(2020)Gururangan, Marasovi{\'c}, Swayamdipta, Lo, Beltagy, Downey, and Smith}]{gururangan-etal-2020-dont}
Suchin Gururangan, Ana Marasovi{\'c}, Swabha Swayamdipta, Kyle Lo, Iz~Beltagy, Doug Downey, and Noah~A. Smith. 2020.
\newblock \href {https://doi.org/10.18653/v1/2020.acl-main.740} {Don{'}t stop pretraining: Adapt language models to domains and tasks}.
\newblock In \emph{Proceedings of the 58th Annual Meeting of the Association for Computational Linguistics}, pages 8342--8360, Online. Association for Computational Linguistics.

\bibitem[{Han et~al.(2024)Han, Ueda, Otake, Katsumata, Kamata, Kiyomaru, Kodama, Sugawara, Chen, Matsuda, Miyao, Miyawaki, and Ryu}]{han-kiyomaru-etal-2024-llm-jp-eval}
Namgi Han, Nobuhiro Ueda, Masatoshi Otake, Satoru Katsumata, Keisuke Kamata, Hirokazu Kiyomaru, Takashi Kodama, Saku Sugawara, Bowen Chen, Hiroshi Matsuda, Yusuke Miyao, Yugo Miyawaki, and Koki Ryu. 2024.
\newblock Automatic evaluation tool for {J}apanese large language models [llm-jp-eval: 日本語大規模言語モデルの自動評価ツール] (in {J}apanese).
\newblock In \emph{NLP 2024}.

\bibitem[{Han and Eisenstein(2019)}]{han-eisenstein-2019-unsupervised}
Xiaochuang Han and Jacob Eisenstein. 2019.
\newblock \href {https://doi.org/10.18653/v1/D19-1433} {Unsupervised domain adaptation of contextualized embeddings for sequence labeling}.
\newblock In \emph{Proceedings of the 2019 Conference on Empirical Methods in Natural Language Processing and the 9th International Joint Conference on Natural Language Processing (EMNLP-IJCNLP)}, pages 4238--4248, Hong Kong, China. Association for Computational Linguistics.

\bibitem[{Huang et~al.(2023)Huang, Tang, Zhang, Zhao, Song, Xia, and Wei}]{huang-etal-2023-languages}
Haoyang Huang, Tianyi Tang, Dongdong Zhang, Xin Zhao, Ting Song, Yan Xia, and Furu Wei. 2023.
\newblock \href {https://doi.org/10.18653/v1/2023.findings-emnlp.826} {Not all languages are created equal in {LLM}s: Improving multilingual capability by cross-lingual-thought prompting}.
\newblock In \emph{Findings of the Association for Computational Linguistics: EMNLP 2023}, pages 12365--12394, Singapore. Association for Computational Linguistics.

\bibitem[{Ishii et~al.(2023)Ishii, Inoue, and Sekine}]{ishii-jemhopqa}
Ai~Ishii, Naoya Inoue, and Satoshi Sekine. 2023.
\newblock \href {https://www.anlp.jp/proceedings/annual_meeting/2023/pdf_dir/Q8-14.pdf} {Construction of a {J}apanese multi-hop {QA} dataset for a question-answering system that can explain its reasons [根拠を説明可能な質問応答システムのための日本語マルチホップ{QA}データセット構築] (in {J}apanese)}.
\newblock In \emph{NLP 2023}.

\bibitem[{Jiang et~al.(2022)Jiang, Liang, Chen, and Duan}]{XLM-K-2022-aaai}
Xiaoze Jiang, Yaobo Liang, Weizhu Chen, and Nan Duan. 2022.
\newblock \href {https://arxiv.org/abs/2109.12573} {{XLM-K}: Improving cross-lingual language model pre-training with multilingual knowledge}.
\newblock In \emph{Proceedings of the Thirty-Sixth AAAI Conference on Artificial Intelligence}.

\bibitem[{Joshi et~al.(2020)Joshi, Santy, Budhiraja, Bali, and Choudhury}]{joshi-etal-2020-state}
Pratik Joshi, Sebastin Santy, Amar Budhiraja, Kalika Bali, and Monojit Choudhury. 2020.
\newblock \href {https://doi.org/10.18653/v1/2020.acl-main.560} {The state and fate of linguistic diversity and inclusion in the {NLP} world}.
\newblock In \emph{Proceedings of the 58th Annual Meeting of the Association for Computational Linguistics}, pages 6282--6293, Online. Association for Computational Linguistics.

\bibitem[{Kale et~al.(2021)Kale, Siddhant, Al-Rfou, Xue, Constant, and Johnson}]{kale-etal-2021-nmt5}
Mihir Kale, Aditya Siddhant, Rami Al-Rfou, Linting Xue, Noah Constant, and Melvin Johnson. 2021.
\newblock \href {https://doi.org/10.18653/v1/2021.acl-short.87} {nm{T}5 - is parallel data still relevant for pre-training massively multilingual language models?}
\newblock In \emph{Proceedings of the 59th Annual Meeting of the Association for Computational Linguistics and the 11th International Joint Conference on Natural Language Processing (Volume 2: Short Papers)}, pages 683--691, Online. Association for Computational Linguistics.

\bibitem[{Kurihara et~al.(2022)Kurihara, Kawahara, and Shibata}]{kurihara-etal-2022-jglue}
Kentaro Kurihara, Daisuke Kawahara, and Tomohide Shibata. 2022.
\newblock \href {https://aclanthology.org/2022.lrec-1.317} {{JGLUE}: {J}apanese general language understanding evaluation}.
\newblock In \emph{Proceedings of the Thirteenth Language Resources and Evaluation Conference}, pages 2957--2966, Marseille, France. European Language Resources Association.

\bibitem[{Lin et~al.(2021)Lin, Lee, Qiao, and Ren}]{lin-etal-2021-common}
Bill~Yuchen Lin, Seyeon Lee, Xiaoyang Qiao, and Xiang Ren. 2021.
\newblock \href {https://doi.org/10.18653/v1/2021.acl-long.102} {Common sense beyond {E}nglish: Evaluating and improving multilingual language models for commonsense reasoning}.
\newblock In \emph{Proceedings of the 59th Annual Meeting of the Association for Computational Linguistics and the 11th International Joint Conference on Natural Language Processing (Volume 1: Long Papers)}, pages 1274--1287, Online. Association for Computational Linguistics.

\bibitem[{Loshchilov and Hutter(2017)}]{Loshchilov2017FixingWD}
Ilya Loshchilov and Frank Hutter. 2017.
\newblock \href {https://api.semanticscholar.org/CorpusID:3312944} {Fixing weight decay regularization in {A}dam}.
\newblock \emph{ArXiv}, abs/1711.05101.

\bibitem[{M{\"u}ller and Laurent(2022)}]{muller2022cedille}
Martin M{\"u}ller and Florian Laurent. 2022.
\newblock \href {https://api.semanticscholar.org/CorpusID:246634768} {Cedille: A large autoregressive {F}rench language model}.
\newblock \emph{ArXiv}, abs/2202.03371.

\bibitem[{Neubig and Hu(2018)}]{neubig-hu-2018-rapid}
Graham Neubig and Junjie Hu. 2018.
\newblock \href {https://doi.org/10.18653/v1/D18-1103} {Rapid adaptation of neural machine translation to new languages}.
\newblock In \emph{Proceedings of the 2018 Conference on Empirical Methods in Natural Language Processing}, pages 875--880, Brussels, Belgium. Association for Computational Linguistics.

\bibitem[{Nguyen et~al.(2023)Nguyen, Zhang, Li, Aljunied, Tan, Cheng, Chen, Deng, Yang, Liu, Zhang, and Bing}]{nguyen2023seallms}
Xuan-Phi Nguyen, Wenxuan Zhang, Xin Li, Mahani Aljunied, Qingyu Tan, Liying Cheng, Guanzheng Chen, Yue Deng, Sen Yang, Chaoqun Liu, Hang Zhang, and Li~Bing. 2023.
\newblock \href {https://api.semanticscholar.org/CorpusID:265551745} {{SeaLLMs} - large language models for {S}outheast {A}sia}.
\newblock \emph{ArXiv}, abs/2312.00738.

\bibitem[{Rajbhandari et~al.(2019)Rajbhandari, Rasley, Ruwase, and He}]{Rajbhandari2019ZeROMO}
Samyam Rajbhandari, Jeff Rasley, Olatunji Ruwase, and Yuxiong He. 2019.
\newblock \href {https://api.semanticscholar.org/CorpusID:203736482} {{ZeRO}: Memory optimizations toward training trillion parameter models}.
\newblock \emph{SC20: International Conference for High Performance Computing, Networking, Storage and Analysis}, pages 1--16.

\bibitem[{Reid and Artetxe(2022)}]{reid-artetxe-2022-paradise}
Machel Reid and Mikel Artetxe. 2022.
\newblock \href {https://doi.org/10.18653/v1/2022.naacl-main.58} {{PARADISE}: Exploiting parallel data for multilingual sequence-to-sequence pretraining}.
\newblock In \emph{Proceedings of the 2022 Conference of the North American Chapter of the Association for Computational Linguistics: Human Language Technologies}, pages 800--810, Seattle, United States. Association for Computational Linguistics.

\bibitem[{Ri et~al.(2022)Ri, Yamada, and Tsuruoka}]{ri-etal-2022-mluke}
Ryokan Ri, Ikuya Yamada, and Yoshimasa Tsuruoka. 2022.
\newblock \href {https://doi.org/10.18653/v1/2022.acl-long.505} {m{LUKE}: {T}he power of entity representations in multilingual pretrained language models}.
\newblock In \emph{Proceedings of the 60th Annual Meeting of the Association for Computational Linguistics (Volume 1: Long Papers)}, pages 7316--7330, Dublin, Ireland. Association for Computational Linguistics.

\bibitem[{Sekine(2003)}]{sekine-niilc}
Satoshi Sekine. 2003.
\newblock \href {https://www.anlp.jp/proceedings/annual_meeting/2003/pdf_dir/C7-6.pdf} {Development of a question answering system focused on an encyclopedia [百科事典を対象とした質問応答システムの開発] (in {J}apanese)}.
\newblock In \emph{NLP 2003}.

\bibitem[{Suzuki et~al.(2020)Suzuki, Suzuki, Matsuda, Nishida, and Inoue}]{suzuki-jaqket}
Masatoshi Suzuki, Jun Suzuki, Koji Matsuda, Kyosuke Nishida, and Naoya Inoue. 2020.
\newblock \href {https://www.anlp.jp/proceedings/annual_meeting/2020/pdf_dir/P2-24.pdf} {{JAQKET}: Constructing a {J}apanese {QA} dataset based on quiz questions [{JAQKET}:クイズを題材にした日本語{QA}データセットの構築] (in {J}apanese)}.
\newblock In \emph{NLP 2020}.

\bibitem[{Talmor et~al.(2019)Talmor, Herzig, Lourie, and Berant}]{talmor-etal-2019-commonsenseqa}
Alon Talmor, Jonathan Herzig, Nicholas Lourie, and Jonathan Berant. 2019.
\newblock \href {https://doi.org/10.18653/v1/N19-1421} {{C}ommonsense{QA}: A question answering challenge targeting commonsense knowledge}.
\newblock In \emph{Proceedings of the 2019 Conference of the North {A}merican Chapter of the Association for Computational Linguistics: Human Language Technologies, Volume 1 (Long and Short Papers)}, pages 4149--4158, Minneapolis, Minnesota. Association for Computational Linguistics.

\bibitem[{Touvron et~al.(2023)Touvron, Martin, Stone, Albert, Almahairi, Babaei, Bashlykov, Batra, Bhargava, Bhosale, Bikel, Blecher, Ferrer, Chen, Cucurull, Esiobu, Fernandes, Fu, Fu, Fuller, Gao, Goswami, Goyal, Hartshorn, Hosseini, Hou, Inan, Kardas, Kerkez, Khabsa, Kloumann, Korenev, Koura, Lachaux, Lavril, Lee, Liskovich, Lu, Mao, Martinet, Mihaylov, Mishra, Molybog, Nie, Poulton, Reizenstein, Rungta, Saladi, Schelten, Silva, Smith, Subramanian, Tan, Tang, Taylor, Williams, Kuan, Xu, Yan, Zarov, Zhang, Fan, Kambadur, Narang, Rodriguez, Stojnic, Edunov, and Scialom}]{touvron2023llama}
Hugo Touvron, Louis Martin, Kevin~R. Stone, Peter Albert, Amjad Almahairi, Yasmine Babaei, Nikolay Bashlykov, Soumya Batra, Prajjwal Bhargava, Shruti Bhosale, Daniel~M. Bikel, Lukas Blecher, Cristian~Cant{\'o}n Ferrer, Moya Chen, Guillem Cucurull, David Esiobu, Jude Fernandes, Jeremy Fu, Wenyin Fu, Brian Fuller, Cynthia Gao, Vedanuj Goswami, Naman Goyal, Anthony~S. Hartshorn, Saghar Hosseini, Rui Hou, Hakan Inan, Marcin Kardas, Viktor Kerkez, Madian Khabsa, Isabel~M. Kloumann, A.~V. Korenev, Punit~Singh Koura, Marie-Anne Lachaux, Thibaut Lavril, Jenya Lee, Diana Liskovich, Yinghai Lu, Yuning Mao, Xavier Martinet, Todor Mihaylov, Pushkar Mishra, Igor Molybog, Yixin Nie, Andrew Poulton, Jeremy Reizenstein, Rashi Rungta, Kalyan Saladi, Alan Schelten, Ruan Silva, Eric~Michael Smith, R.~Subramanian, Xia Tan, Binh Tang, Ross Taylor, Adina Williams, Jian~Xiang Kuan, Puxin Xu, Zhengxu Yan, Iliyan Zarov, Yuchen Zhang, Angela Fan, Melanie Kambadur, Sharan Narang, Aurelien Rodriguez, Robert Stojnic, Sergey Edunov, and
  Thomas Scialom. 2023.
\newblock \href {https://api.semanticscholar.org/CorpusID:259950998} {Llama 2: Open foundation and fine-tuned chat models}.
\newblock \emph{ArXiv}, abs/2307.09288.

\bibitem[{Wang et~al.(2022)Wang, Ruder, and Neubig}]{wang-etal-2022-expanding}
Xinyi Wang, Sebastian Ruder, and Graham Neubig. 2022.
\newblock \href {https://doi.org/10.18653/v1/2022.acl-long.61} {Expanding pretrained models to thousands more languages via lexicon-based adaptation}.
\newblock In \emph{Proceedings of the 60th Annual Meeting of the Association for Computational Linguistics (Volume 1: Long Papers)}, pages 863--877, Dublin, Ireland. Association for Computational Linguistics.

\bibitem[{Wang et~al.(2020)Wang, K, Mayhew, and Roth}]{wang-etal-2020-extending}
Zihan Wang, Karthikeyan K, Stephen Mayhew, and Dan Roth. 2020.
\newblock \href {https://doi.org/10.18653/v1/2020.findings-emnlp.240} {Extending multilingual {BERT} to low-resource languages}.
\newblock In \emph{Findings of the Association for Computational Linguistics: EMNLP 2020}, pages 2649--2656, Online. Association for Computational Linguistics.

\bibitem[{Yong et~al.(2023)Yong, Schoelkopf, Muennighoff, Aji, Adelani, Almubarak, Bari, Sutawika, Kasai, Baruwa, Winata, Biderman, Raff, Radev, and Nikoulina}]{yong-etal-2023-bloom}
Zheng~Xin Yong, Hailey Schoelkopf, Niklas Muennighoff, Alham~Fikri Aji, David~Ifeoluwa Adelani, Khalid Almubarak, M~Saiful Bari, Lintang Sutawika, Jungo Kasai, Ahmed Baruwa, Genta Winata, Stella Biderman, Edward Raff, Dragomir Radev, and Vassilina Nikoulina. 2023.
\newblock \href {https://doi.org/10.18653/v1/2023.acl-long.653} {{BLOOM}+1: Adding language support to {BLOOM} for zero-shot prompting}.
\newblock In \emph{Proceedings of the 61st Annual Meeting of the Association for Computational Linguistics (Volume 1: Long Papers)}, pages 11682--11703, Toronto, Canada. Association for Computational Linguistics.

\bibitem[{Zhao et~al.(2024)Zhao, Zhang, Zhang, Gui, and Huang}]{zhao2024llama}
Jun Zhao, Zhihao Zhang, Qi~Zhang, Tao Gui, and Xuanjing Huang. 2024.
\newblock \href {https://api.semanticscholar.org/CorpusID:266725709} {{LLaMA} beyond {E}nglish: An empirical study on language capability transfer}.
\newblock \emph{ArXiv}, abs/2401.01055.

\end{thebibliography}
\bibliographystyle{acl_natbib}

\appendix

\clearpage

\section*{Appendix for ``LEIA: Facilitating Cross-lingual Knowledge Transfer in Language Models with Entity-based Data Augmentation''}

\section{Details of Training}
\label{sec:detail-training}

\head{Preprocessing} The training corpus is derived from the target language edition of October 2023 Wikipedia dump.\footnote{\url{https://dumps.wikimedia.org/}}
We extract text and hyperlinks using the open-source WikiExtractor tool.\footnote{\url{https://github.com/attardi/wikiextractor}} Each entity referenced by a hyperlink is mapped to its English equivalent using the inter-language link database obtained from the October 2023 Wikidata dump.\footnote{\url{https://dumps.wikimedia.org/wikidatawiki/}} We filter out entities whose English names begin with the following prefixes, denoting special Wikipedia entities:
\begin{itemize}
\itemsep -0.1em 
\item Book:
\item Category:
\item Draft:
\item File:
\item Help:
\item List of 
\item MediaWiki:
\item Portal:
\item Special:
\item Talk:
\item Template:
\item User:
\item Wikipedia:
\item WikiProject:
\end{itemize}
Additionally, we remove suffix strings enclosed in parentheses from entity names, e.g., ``(state)'' in ``Washington (state)''.
The entity names are enclosed in the special \textit{<translate>} and \textit{</translate>} tokens, and inserted into the Wikipedia text as described in §\ref{sec:method}.

\head{Training}
We train our models using the AdamW optimizer~\cite{Loshchilov2017FixingWD} with $\beta_1=0.99$ and $\beta_2=0.95$, following \citet{touvron2023llama}.
We use a cosine learning rate schedule with an initial learning rate of 5e-6.
The model checkpoint corresponding to the last training step is used as the final model.
The detailed hyperparameters are shown in Table \ref{tab:hyperparams}.
To enhance training efficiency, we create an input sequence by concatenating multiple short Wikipedia articles until the maximum token length is reached.

\begin{table}[t]
    \centering
    \begin{tabular}{cc}
        \toprule
        Name & Value\\
        \midrule
        Max token length & 2,048 \\
        Batch size & 2,048 \\
        Optimizer & AdamW \\
        Adam $\beta_1$ & 0.9 \\
        Adam $\beta_2$ & 0.95 \\
        Initial learning rate & 5e-6 \\
        Learning rate schedule & Cosine \\
        Warmup steps & 0 \\
        Max gradient norm & 1.0 \\
        Weight decay & 0.1 \\
        \bottomrule
    \end{tabular}
    \caption{Hyperparameters used to fine-tune the model.}
    \label{tab:hyperparams}
\end{table}

The special tokens, \textit{i.e.,} \textit{<translate>} and \textit{</translate>}, are added to the vocabulary of the model and their embeddings are initialized using the mean embedding derived from all token embeddings.

We use a machine with eight Nvidia A100 40GB to train the model.
The LLaMA 2 \cite{touvron2023llama} and Swallow \cite{fuji-nakamura-etal-2024-swallow-llm} models, both available under the LLaMA 2 Community License\footnote{\url{https://ai.meta.com/llama/license/}}, are used as our base models. 
The training approximately takes 2 hours for Swahili in the experiments with LLaMA 2 and 5 hours for the other models.
We adopt data parallelism with sharding parameters across GPUs using DeepSpeed Zero Redundancy Optimizer~\cite{Rajbhandari2019ZeROMO}, mixed precision training with bfloat16, and FlashAttention-2~\cite{Dao2023FlashAttention2FA} to reduce computational costs.

\begin{table}
    \centering
    \scalebox{0.94}{
    \begin{tabular}{lll}
        \toprule
        Code & Name & Family\\
        \midrule
        ar & Arabic & Afro-Asiatic \\
        es & Spanish & Indo-European (Italic) \\
        hi & Hindi & Indo-European (Indo-Iranian)\\
        ja & Japanese &  Japonic \\
        ru & Russian & Indo-European (Balto-Slavic) \\
        sw & Swahili & Niger-Congo \\
        zh & Chinese & Sino-Tibetan\\
        \bottomrule
    \end{tabular}
    }
    \caption{List of languages with their codes and families used in our experiments.}
    \label{tab:languages}
\end{table}

\begin{table*}[t]
    \centering
    \setlength{\tabcolsep}{0.5em}
    \scalebox{0.81}{
    \begin{tabular}{cc|cccccccc|cccccccc}
        \toprule
        & &  \multicolumn{8}{c|}{X-CODAH} & \multicolumn{8}{c}{X-CSQA} \\
        strategy & $p_\text{skip} $ & ar & es & hi & ja & ru & sw & zh & avg & ar & es & hi & ja & ru & sw & zh & avg \\
        \midrule
        left & 0.0 & 32.0 & 46.7 & 30.3 & 34.0 & 38.3 & 30.0 & 38.0 & 35.6 & 21.9 & 46.0 & 19.3 & 34.6 & 35.6 & 16.2 & 39.9 & 30.5 \\
        left & 0.5 & 33.0 & 46.7 & 30.3 & 35.0 & 37.7 & 30.7 & 39.3 & 36.1 & 22.0 & 46.0 & 18.4 & 35.3 & 35.8 & 15.9 & 40.5 & 30.6 \\
        \midrule
        right & 0.0 & 32.3 & 46.7 & 30.3 & 34.3 & 38.3 & 30.3 & 38.3 & 35.8 & 21.8 & 45.6 & 19.3 & 34.6 & 35.8 & 16.4 & 39.9 & 30.5 \\
        right & 0.5 & 33.3 & 46.7 & 30.7 & 35.0 & 37.7 & 30.3 & 39.3 & 36.1 & 22.0 & 45.8 & 18.3 & 35.5 & 36.0 & 16.1 & 40.4 & 30.6 \\
        \midrule
        replace & 0.0 & 32.0 & 46.7 & 30.0 & 34.7 & 38.0 & 30.0 & 39.0 & 35.8 & 21.8 & 45.5 & 18.7 & 34.6 & 35.6 & 16.3 & 40.1 & 30.4 \\
        replace & 0.5 & 33.0 & 46.3 & 30.3 & 35.3 & 37.0 & 30.7 & 39.0 & 36.0 & 21.8 & 45.8 & 18.4 & 35.5 & 36.0 & 15.9 & 40.3 & 30.5 \\
        \bottomrule

    \end{tabular}
    }
    \caption{Detailed accuracy scores across seven languages based on different method configurations. Average scores correspond to those in Table \ref{tab:avg-scores-all-configurations}.}
    \label{tab:full-scores-all-configurations}
\end{table*}

\begin{table*}[t]
    \centering
    \setlength{\tabcolsep}{0.4em}
    \scalebox{0.65}{
    \begin{tabular}{p{28em}|p{34em}}
        \toprule
        LLaMA2-FT's predictions & LEIA's predictions\\
        \midrule
        海を沸かせようとしている。 海が大好きなんです。 & 海を沸かせようとしている。 彼は不可能なことを成し遂げようとしている。 \\
        イギリスのロンドンにあるエッフェル塔。 & エッフェル塔はパリにあります。 \\
        電話が鳴る音がする。 音量を上げてみました。 & 電話が鳴る音がする。 私はそれに答える。 \\
        お酒を飲みすぎてしまいます。 小学4年生になりました。 & お酒を飲みすぎてしまいます。 私はアルコール依存症かもしれません。 \\
        フードコートに人がいる。 その人はマラソンを走っています。 & フードコートに人がいる。 その人はサンドイッチを買う。 \\
         \bottomrule
    \end{tabular}
    }
    \caption{Comparison of X-CODAH predictions by LLaMA2-FT and LEIA. Japanese sentences used in experiments are shown here. See Table \ref{tab:qualitative-analysis} for corresponding English sentences.}
    \label{tab:qualitative-analysis-japanese}
\end{table*}

\begin{table*}[t]
    \centering
    \setlength{\tabcolsep}{0.5em}
    \scalebox{0.74}{
    \begin{tabular}{c|cccccccc|cccccccc}
        \toprule
        & \multicolumn{8}{c|}{X-CODAH} & \multicolumn{8}{c}{X-CSQA} \\
         & ar & es & hi & ja & ru & sw & zh & avg & ar & es & hi & ja & ru & sw & zh & avg \\
        \midrule
        \begin{tabular}{@{}c@{}}Enable loss propagation\\from English entity tokens\end{tabular} & 33.3 & 46.7 & 30.7 & 35.0 & 37.7 & 30.3 & 39.3 & 36.1 & 22.0 & 45.8 & 18.3 & 35.5 & 36.0 & 16.1 & 40.4 & 30.6 \\
        \begin{tabular}{@{}c@{}}Disable loss propagation\\from English entity tokens\end{tabular} & 32.3 & 46.3 & 30.3 & 35.3 & 37.7 & 30.7 & 39.0 & 36.0 & 21.6 & 46.1 & 18.5 & 35.5 & 36.1 & 16.1 & 40.3 & 30.6 \\
        \bottomrule
    \end{tabular}
    }
    \caption{Comparison of LEIA models with loss propagation enabled vs. disabled for tokens in English entity names. Average scores correspond to those in Table \ref{tab:prevent-token-loss}.}
    \label{tab:prevent-token-loss-full-results}
\end{table*}

\begin{table}[t]
    \centering
    \scalebox{0.93}{
    \begin{tabular}{c|cc|cc}
        \toprule
        & \multicolumn{2}{c|}{X-CODAH} & \multicolumn{2}{c}{X-CSQA} \\
        Model & 4-shot & 7-shot & 4-shot & 7-shot \\
        \midrule
        LLaMA2 & 33.6 & 33.6 & 30.2 & 30.5 \\
        LLaMA2+FT & 33.3 & 33.3 & 29.9 & 30.4 \\
        \midrule
        LEIA & \highestscore{36.1} & \highestscore{36.1} & \highestscore{30.6} & \highestscore{30.9} \\
        \bottomrule
    \end{tabular}
    }
    \caption{Average accuracy scores across seven languages based on four- and seven-shot examples.}
    \label{tab:avg-scores-varied-few-shots}
\end{table}

\section{Details of Languages Used in LLaMA 2 Experiments}
\label{sec:languages}

Our experiments with LLaMA 2 (§\ref{sec:experiments_llama2}) are conducted in seven languages from five diverse language families shown in Table \ref{tab:languages}.

\section{Full Experimental Results}

The detailed per-language results of comparing different method configurations of LEIA are presented in Table \ref{tab:full-scores-all-configurations}. 
The Japanese sentences corresponding to the English sentences in Table \ref{tab:qualitative-analysis} can be found in Table \ref{tab:qualitative-analysis-japanese}.
The detailed results for our models with loss propagation enabled and disabled when predicting tokens in English entity names are provided in Table \ref{tab:prevent-token-loss-full-results}.

\section{Results with Varied Number of Few-shot Examples}
\label{sec:various-num-of-few-shots}

The experimental results with LLaMA 2 using four- and seven-shot examples are available in Table \ref{tab:avg-scores-varied-few-shots}.
LEIA consistently outperforms the baseline models in both settings.

\section{Details of Experiments}
\label{sec:detail-experiments}

We evaluate our models using two multilingual question answering datasets, X-CODAH and X-CSQA~\cite{lin-etal-2021-common}, along with four Japanese question answering datasets: JEMHopQA, NIILC, JCommonsenseQA, and JAQKET.
While inputs of X-CODAH consist of single texts, inputs for the other datasets are divided into questions and answers.
This necessitates specifying the input format for the model, leading us to use a few-shot setting for datasets other than X-CODAH, instead of a zero-shot setting.

For the JEMHopQA and NIILC datasets, which do not provide answer candidates, the model generates textual answers, and its performance is measured using a character-based F-measure, following \citet{han-kiyomaru-etal-2024-llm-jp-eval}.
For other tasks, we input each answer candidate into the models, and select the one with the highest probability.
We use the llm-jp-eval tool for JEMHopQA and NIILC, and the JP Language Model Evaluation Harness for JCommonsenseQA and JAQKET.
The prompts for our experiments are presented in Figures \ref{fig:prompt-xcsqa}–\ref{fig:prompt-jaqket}, with \textit{\{question\}}, \textit{\{answer\}}, and \textit{\{choiceX\}} replaced by the actual question, answer, and answer candidates, respectively.
The prompts shown in Figures \ref{fig:prompt-jemhopqa}–\ref{fig:prompt-niilc} and Figures \ref{fig:prompt-jcommonsenseqa}–\ref{fig:prompt-jaqket} are the default prompts in llm-jp-eval and JP Language Model Evaluation Harness, respectively.

The detailed descriptions of the datasets used in our experiments are provided as follows:
\begin{figure}[t]
\makebox[\columnwidth][c]{
\scalebox{0.9}{
\fbox{
\begin{minipage}{\columnwidth}
Question: \textit{\{question\}}\\
Answer: \textit{\{answer\}}
\end{minipage}
}}}
\caption{Prompt for X-CSQA.}
\label{fig:prompt-xcsqa}
\end{figure}

\begin{figure}[t]
\makebox[\columnwidth][c]{
\scalebox{0.9}{
\fbox{
\begin{minipage}{\columnwidth}
以下はタスクを説明する指示と、追加の背景情報を提供する入力の組み合わせです。要求を適切に満たす回答を書いてください。\\
\#\#\# 指示\\
質問を入力とし、回答を出力してください。回答の他には何も含めないことを厳守してください。\\
\\
\#\#\# 入力：\\
\textit{\{question\}}\\
\#\#\# 回答：\\
\textit{\{answer\}}
  \end{minipage}
}}}
\caption{Prompt for JEMHopQA in llm-jp-eval.}
\label{fig:prompt-jemhopqa}
\end{figure}

\begin{figure}[t]
\makebox[\columnwidth][c]{
\scalebox{0.9}{
\fbox{
\begin{minipage}{\columnwidth}
以下はタスクを説明する指示と、追加の背景情報を提供する入力の組み合わせです。要求を適切に満たす回答を書いてください。\\
\#\#\# 指示\\
質問に対する答えを出力してください。答えが複数の場合、コンマ（,）で繋げてください。\\
\\
\#\#\# 入力：\\
\textit{\{question\}}\\
\#\#\# 回答：\\
\textit{\{answer\}}
  \end{minipage}
}}}
\caption{Prompt for NIILC in llm-jp-eval.}
\label{fig:prompt-niilc}
\end{figure}

\begin{figure}[t]
\makebox[\columnwidth][c]{
\scalebox{0.9}{
\fbox{
\begin{minipage}{\columnwidth}
与えられた選択肢の中から、最適な答えを選んでください。\\
\\
質問：\textit{\{question\}}\\
選択肢：\\
- \textit{\{choice0\}}\\
- \textit{\{choice1\}}\\
- \textit{\{choice2\}}\\
- \textit{\{choice3\}}\\
- \textit{\{choice4\}}\\
回答：\textit{\{answer\}}
  \end{minipage}
}}}
\caption{Prompt for JCommonsenseQA in JP Language Model Evaluation Harness.}
\label{fig:prompt-jcommonsenseqa}
\end{figure}

\begin{figure}[t]
\makebox[\columnwidth][c]{
\scalebox{0.9}{
\fbox{
\begin{minipage}{\columnwidth}
文章と質問と回答の選択肢を入力として受け取り、選択肢から質問に対する回答を選択してください。なお、回答は選択肢の番号(例:0)でするものとします。\\
\\
質問：\textit{\{question\}}\\
選択肢:0.\textit{\{choice0\}},1.\textit{\{choice1\}},1.\textit{\{choice1\}},\\2.\textit{\{choice2\}},3.\textit{\{choice3\}},4.\textit{\{choice4\}}\\
回答:\textit{\{answer\}}
  \end{minipage}
}}}
\caption{Prompt for JAQKET in JP Language Model Evaluation Harness.}
\label{fig:prompt-jaqket}
\end{figure}

\begin{itemize}
\itemsep -0.1em 
\item \textbf{X-CODAH}~\cite{lin-etal-2021-common} is a four-way multiple-choice, multilingual question answering dataset created by translating the English CODAH dataset~\cite{chen-etal-2019-codah}.
We use the validation set, consisting of 300 examples, as the test set labels are not publicly accessible.
We do not use a hand-crafted prompt and simply input the original text.
This dataset is obtained from the corresponding Hugging Face repository\footnote{\url{https://huggingface.co/datasets/xcsr}} licensed under the MIT license.

\item \textbf{X-CSQA}~\cite{lin-etal-2021-common} is a five-way multiple-choice, multilingual question answering dataset, translated from the CommonsenseQA dataset~\cite{talmor-etal-2019-commonsenseqa}.
Due to the unavailability of test set labels, we use the validation set, which comprises 1,000 instances.
Few-shot examples are randomly selected from the same set. 
The prompt for this dataset is shown in Figure \ref{fig:prompt-xcsqa}.
This dataset is obtained from the same repository as the X-CODAH dataset.

\item \textbf{JEMHopQA}~\cite{ishii-jemhopqa} is a Japanese question answering dataset with 120 input questions and their ideal answers.
Few-shot examples are selected from its dedicated set.
The prompt for this dataset is shown in Figure \ref{fig:prompt-jemhopqa}.
This dataset is licensed under the Creative Commons CC BY-SA 4.0 license.

\item \textbf{NIILC}~\cite{sekine-niilc} is a Japanese question answering dataset comprising 198 input questions and their ideal answers.
Few-shot examples are selected from its dedicated set.
The prompt for this dataset is shown in Figure \ref{fig:prompt-niilc}.
This dataset is licensed under the Creative Commons CC BY-SA 4.0 license.

\item \textbf{JCommonsenseQA}~\cite{kurihara-etal-2022-jglue} is a Japanese five-way multiple-choice question answering dataset.
We use 1,119 examples from the validation set.
Few-shot examples are randomly selected from the training set, which comprises 8,939 examples.
The prompt for this dataset is shown in Figure \ref{fig:prompt-jcommonsenseqa}.
This dataset is licensed under the Creative Commons CC BY-SA 4.0 license.

\item \textbf{JAQKET}~\cite{suzuki-jaqket}  is a Japanese five-way multiple-choice question answering dataset.
We use 271 examples from the validation set.
Few-shot examples are randomly selected from the training set, comprising 13,061 examples.
The prompt for this dataset is shown in Figure \ref{fig:prompt-jaqket}.
This dataset is licensed under the Creative Commons CC BY-SA 4.0 license.
\end{itemize}
\end{document}